\pdfoutput=1

\documentclass[11pt]{article}

\usepackage{EMNLP2022}

\usepackage{times}
\usepackage{graphicx}
\usepackage{latexsym}
\usepackage{hyperref}
\usepackage{tabularx}
\usepackage{booktabs}

\usepackage{xargs} 
\usepackage[colorinlistoftodos,prependcaption,textsize=tiny]{todonotes} 

\newcommandx{\zarah}[2][1=]{\todo[linecolor=red,backgroundcolor=red!25,bordercolor=red,#1]{#2}}

\usepackage[T1]{fontenc}

\usepackage[utf8]{inputenc}

\usepackage{microtype}

\usepackage{inconsolata}

\title{How Much Hate with \#china? A Preliminary Analysis on China-related Hateful Tweets Two Years After the Covid Pandemic Began}

\author{Jinghua Xu \\ Heidelberg University\\  jinghua.xu@stud.uni-heidelberg.de
        \And 
        Zarah Weiss \\ University of Tübingen \\weiss.zarah@gmail.com}

\begin{document}
\maketitle
\begin{abstract}
Following the outbreak of a global pandemic, online content is filled with hate speech. Donald Trump's  ``Chinese Virus'' tweet shifted the blame for the spread of the Covid-19 virus to China and the Chinese people, which triggered a new round of anti-China hate both online and offline. This research intends to examine China-related hate speech on Twitter during the two years following the burst of the pandemic (2020 and 2021). Through Twitter's API, in total 2,172,333 tweets hashtagged \#china posted during the time were collected. 

By employing multiple state-of-the-art pretrained language models for hate speech detection, \textcolor{black}{we identify} a wide range of hate of various types, \textcolor{black}{resulting in} an automatically labeled anti-China hate speech dataset. \textcolor{black}{We identify a hateful rate in \#china tweets of 2.5\% in 2020 and 1.9\% in 2021. This is well}  above the average rate of online hate speech on Twitter at 0.6\% \citep{gao2017recognizing}. \textcolor{black}{We further analyzed the longitudinal development of \#china tweets and those identified as hateful in 2020 and 2021 through visualizing the daily number and hate rate over the two years. Our keyword analysis of hate speech in \#china tweets reveals the most frequently mentioned terms in the hateful \#china tweets, which can be used for further social science studies.}

\end{abstract}


\section{Introduction}

\noindent \textcolor{black}{Social media has been identified as a central data resource for computational social sciences \citep[e.g.,][]{oboler2012danger, edelmann2020computational, lazer2009life} and a broad range of applications in computational linguistics \citep[e.g.,][]{potthast-etal-2018-crowdsourcing, corvey2010twitter, coppersmith2014quantifying}. It provides access to an enormous amount of longitudinal language data from a uniquely broad cross-section of the world population. Platforms such as Twitter connect communities across the globe and enable them to share news, ideas, and sentiments in real time. However, this also provides a platform for the spread of fake news and hate speech.} 

Hate speech is commonly defined as any communication that belittles a person or a group based on some characteristic such as race, ethnicity, gender, sexual orientation, nationality, religion, or other characteristics \citep{nockleby2000internet}. As the spread of online hate speech continues to grow, the detection of hate speech on social media has gained increasing significance and visibility \citep{schmidt-wiegand-2017-survey}.

The goals to study hate speech detection are manifold. \textcolor{black}{In computational linguistics, researchers have mainly focused on} \textcolor{black}{flagging hateful, prejudicial, or discriminatory social media contents} \textcolor{black}{to enable online platforms to reduce the toxicity of the online environment by removing or contextualizing hateful posts of users \citep[e.g.,][]{pereira2019detecting, 8292838}.} \textcolor{black}{In practice, this is also relevant for companies that must comply with national laws against the public distribution of hate speech.} 
\textcolor{black}{In computational social science, \textcolor{black}{social media data in general \citep[e.g.,][]{mccormick2017using, mejova2015twitter, ledford2020facebook} and hate speech in particular \citep[e.g.,][]{he2021racism, shen2022xing}} have been identified as a valuable resource to gain insights into the biases and sentiments of specific online communities.}
Analyzing online hate \textcolor{black}{targeting} specific topics \textcolor{black}{or populations also} helps to reveal \textcolor{black}{general opinion trends and emerging public sentiments in relation to local or global events in near real time}. Furthermore, it can help establish a linkage between social factors in social science studies. For instance, \citet{kim2021misinformation} links misinformation regarding China and Covid hate speech using the case of anti-Asian hate speech during the Covid-19 pandemic based on observational data.

Following the outbreak of a global pandemic, \textcolor{black}{there was a notable increase in online hate speech \citep{he2021racism}.} With various restrictions against the virus carried out in countries all over the world, widespread disruption was caused in people's normal lives, which led to rising levels of anxiety, stress, and anger \citep{nicola2020emotional}. On March 16, 2020, then US President Donald Trump linked the Covid-19 virus to China and the Chinese people by referring to Covid-19 as ``Chinese Virus'' in a tweet. The tweet shifted the blame for the global pandemic and redirected the anger to China and the Chinese people. \textcolor{black}{This has led to an overall raised level of anti-China sentiment expressed online \citep{shen2022xing}}.
\textcolor{black}{Studies could show an increase in trending hashtags expressing anti-China sentiments---such as \#fuckchina---on Twitter \citep{he2021racism}.}

While previous studies such as \citet{he2021racism} have examined the Covid-related online hate towards the larger Asian community, this research is particularly interested in anti-China hate on Twitter triggered after the beginning of the pandemic. With \citet{kim2021misinformation} establishing the association between Covid misinformation regarding China and online Covid hate speech, this research intends to examine online hate specifically associated with ``\#china''.
We collect all English tweets posted with the hashtag ``\#china'' in 2020 and 2021, and annotate hate speech using an automatic approach. In order to ensure wide coverage of various types of hate (e.g. immigration, nationality, racism, Covid, etc.), we employ an aggressive approach to identify hateful tweets by using multiple pretrained language models for hate speech detection. The state-of-the-art models have proved excellent performance in previous work \citep{he2021racism, mathew2020hatexplain, barbieri2020tweeteval}. This automatic annotation leads to a silver dataset of anti-China hate speech which we make freely available online.\footnote{\textcolor{black}{The code and data of this paper are released at \href{https://github.com/JINHXu/how-much-hate-with-china}{github.com/JINHXu/how-much-hate-with-china}}.}



\textcolor{black}{Based on this data, we found 2.5\% \#china tweets hateful in 2020 and 1.9\% in 2021. This is well} above the average rate of online hate speech \textcolor{black}{that has been reported in previous research} on Twitter at 0.6\% \citep{gao2017recognizing}. \textcolor{black}{We further explored the development of} \#china tweets and hateful \#china tweets posted per day changing over the two years time span, as well as the daily hate rate. Additionally, we conduct content analysis of the identified hateful tweets to reveal the most frequently mentioned keywords and to observe the topic shift from 2020 to 2021. \textcolor{black}{Our analysis reveals increases in \#china tweets and hateful \#china tweets after socio-political events such as Trump's ``Chinese Virus'' tweet, and an overall higher level of \#china hate in Twitter in 2020 compared to 2021.} \textcolor{black}{This highlights the short- and long-term impact of global events and political discourse on group-specific sentiment expressed on social media.} 

\textcolor{black}{The contributions of this paper are summarized as follows:}

\begin{itemize}

    \item \textcolor{black}{We create a large corpus of 2,172,333 English tweets posted with the hashtag ``\#china'' in the two years following the outbreak of the Covid pandemic (2020 and 2021), with automatic annotation of hate speech through a multi-model approach, which uses three state-of-the-art pre-trained language models for hate speech detection to identify various types of hate (e.g. nationality, race, immigration, and Covid) that may be associated with \#china.}

    \item \textcolor{black}{We present an analysis of the longitudinal development of hate speech in the general \#china discourse from early 2020 to late 2021. Our statistical analysis through data visualization and keyword examination lead to findings that could promote further computational social science research.}

    \item \textcolor{black}{We make our code, data, and analysis publicly available to support future research that seeks to reproduce or elaborate on our work. Resources associated with this paper will be made available on GitHub upon acceptance of the paper.}    
    
\end{itemize}

The remainder of this paper is organized as follows. Section~\ref{sec:related} discusses related work on various hate speech datasets and solutions to hate speech detection. Section~\ref{sec:data} briefly describes the raw data collected in this research, and Section~\ref{sec:method} details the method used in this research for hate speech detection. Section~\ref{sec:analysis} presents \textcolor{black}{our analysis before discussing the impact and conclusion of the present work in Section~\ref{sec:conclusion}.}

\section{Related Work}\label{sec:related}

\textcolor{black}{Most hate speech datasets focus on hate speech targeting a specific group or topic.}
For instance, \citet{warner2012detecting} labeled anti-Semitic hate speech from Yahoo!’s newsgroup post and American Jewish Congress’s website; \citet{kwok2013locate} created a balanced dataset of non-hateful and hateful tweets targeting the African community; \citet{burnap2014tweeting} collected hateful tweets related to the murder of Drummer Lee Rigby in 2013; \citet{basile-etal-2019-semeval} proposes a dataset that contains hate speech targeting women or immigrants. In order to create such datasets, the sources of hate speech data are many. These range from user comments on newspaper articles to online social media content from Facebook, Twitter, Reddit, and other platforms. 
The fact that the majority of hate speech datasets are restricted to a specific type of hate or topic is partially due to the sparsity of online hate speech and the method used to collect raw data for manual annotation. In order to create a hate speech dataset, most research starts from filtering data by searching by keywords in order to gather texts more likely to be hateful and conduct manual annotation on the selected texts. 

\textcolor{black}{With the outbreak of the global Covid pandemic and the strains it has put on societies around the globe, a considerable amount of research over the past two years has focused on Covid-related fake news and hate speech across different geographic and online communities \citep{wu2022predicting, ameur2021aracovid19, cotik2020study, bashar2021progressive}.} \textcolor{black}{Several studies have focused on Covid-related hate speech targeting Asian populations on account of the increase in anti-Asian sentiments linked to the pandemic} \citep[e.g.,][]{nghiem2021stop, an2021predicting, he2021racism}.
\textcolor{black}{\citet{nghiem2021stop} and \citet{he2021racism} created anti-Asian, Covid-related hate speech datasets, and proposed models for its detection, while \citet{an2021predicting} investigates the prediction of anti-Asian hateful users during Covid-19.}
\textcolor{black}{Relatively little work has focused on hate speech targeting specifically China}. \textcolor{black}{\citet{shen2022xing} studies anti-China sentiment on social media between 2016 and 2021 using data from Reddit and 4chan.} \textcolor{black}{As far as we are aware, so far there has been no research on anti-China hate speech using Twitter data.}

Methods for hate speech detection include conventional rule-based approaches (e.g. keyword-based detection, sourcing metadata) and data-driven approaches \citep{macavaney2019hate}. The statistical methods include supervised and unsupervised learning approaches, with supervised methods more widely applied. Various supervised learning models have proved good performance on the task in previous work. These models include both classic machine learning models and neural network models. For classic models, Support Vector Machines \citep{noble2006support}, Naive Bayes \citep{webb2010naive}, and Logistic Regression \citep{wright1995logistic} have been most commonly used. A previous study \citet{putri2020comparison} compared some of the classic models for hate speech detection using Indonesian tweet data. Apart from the classic models, neural network models have also been widely used for the task. For instance, \citet{badjatiya2017deep} investigated various deep learning models for hate speech detection using the benchmark dataset proposed in \citet{waseem2016hateful}. Amongst neural network models, long short-term memory \citep[LSTM;][]{hochreiter1997long} models have been most widely applied and presented excellent performance. A number of systems proposed in previous work are based on or partially based on LSTM \citep{gao2017recognizing, bisht2020detection, de2018hate}. In addition to the traditional models, pretrained language models such as BERT \citep{devlin2018bert} and RoBERTa \citep{liu2019roberta} have proved to advance the state of the art on NLP tasks including hate speech detection. These pretrained models have shown superior performance without overfitting, and are currently among the top performers for the task of hate speech detection.

Despite the advanced performance of various supervised learning methods, these approaches require a large amount of annotated data, which are costly to create and often restricted to specific types. Many unsupervised methods for hate speech detection have been developed and used over the years. For instance, \citet{gao2017recognizing} proposed a weakly supervised two-path bootstrapping system, which was designed to capture both implicit and explicit hate speech with the minimum requirement for annotated data. The bootstrapping system contains two learning components: a slur term learner and an LSTM classifier. Due to the low reliability of the slur term learner, the overall system performance is modest. Later work such as ``Snorkel'' proposed in \citet{ratner2020snorkel} is able to achieve better performance with weak supervision by statistically modeling the process of rule-based labeling and training high-accuracy machine learning models for various NLP tasks including text classification.

Against this research background, this paper uses the recently emerged hate indicator keyword \textit{\#china} to collect tweets, and identifies hateful contents utilizing the state-of-the-art pretrained language models, in order to create a new hate speech dataset specifically focusing on anti-China hate, which has gained increasing visibility in online community since the burst of the Covid-19 pandemic yet never been portrayed in a previous study. Furthermore, work prior to ours generally examined the online hate rate on Twitter over a shorter period of time, while our research also intends to find out the percentage of hateful content amongst tweets posted with \#china over the longer two-year time span following the outbreak of the global pandemic.

\section{Data}\label{sec:data}

The data have been collected through Twitter's API. This research collects all English tweets\footnote{Exclude retweets, quotes, and replies.} posted with the hashtag \#china in 2020 and 2021. In 2020, in total 1,236,335 tweets are collected, and 935,998 tweets in 2021. Table \ref{Table 1} presents an overview of the number of \#china tweets in each quarter over the two years. Table \ref{Table 2} summarises the approximate word counts of tweets posted with the hashtag per quarter in the two years.
\begin{table}[h]
\centering
\begin{tabularx}{\columnwidth}{ccccc}
\toprule
Year & Q1 & Q2 & Q3 & Q4 \\
\midrule
2020 & 336,017 & 393,513 & 295,283 & 211,522 \\
2021 & 217,831 & 232,949 & 243,974 & 241,244 \\
\bottomrule
\end{tabularx}
\caption{The number of tweets posted with the hashtag \#china per quarter in 2020 and 2021.}\label{Table 1}
\end{table}

\begin{table}[h]
\centering
\begin{tabularx}{\columnwidth}{ccccc}
\toprule
Year & Q1 & Q2 & Q3 & Q4 \\
\midrule
2020 & 8,521 K & 9,834 K & 7,219 K & 5,199 K \\
2021 & 5,757 K & 5,975 K & 6,342 K & 6,253 K \\
\bottomrule
\end{tabularx}
\caption{The approximate number of words (URLs excluded) in tweets posted with the hashtag \#china per quarter in 2020 and 2021.}\label{Table 2}
\end{table}

In addition to tweet content and creation time, the dataset also includes other metadata of each tweet including author id, tweet id, like count, quote count, reply count, retweet count, and source.

\section{Method}\label{sec:method}
In order to ensure the reliability of the predictions, this research employs three different state-of-the-art pretrained language models for hate speech detection: \textcolor{black}{the COVID-HATE BERT model \citep{he2021racism}, the HateXplain BERT model \citep{mathew2020hatexplain}, and the Twitter RoBERTa Hate model \citep{barbieri2020tweeteval}. The selected models are all trained at least partially on English tweet data. We discuss each model in more detail in the following subsections.}

\textcolor{black}{We employ three different hate speech detection models} to ensure a wide coverage of various types of hate speech \textcolor{black}{and to mitigate model biases \citep{sap-etal-2019-risk}.}
\textcolor{black}{To give an example}, models trained on hate speech targeting the Asian community may be more sensitive \textcolor{black}{to racist hate speech than to sexist hate speech}. 
\textcolor{black}{However, our study is interested in the general increase of hate speech of any type under the hashtag \#china.} \#china hate may be explicitly expressed in various forms targeting different groups of people, for instance, hate towards people with a Chinese nationality (despite their ethnicity group), hate towards Chinese people as a racial group regardless of their nationality or other attributes, and hate towards Chinese immigrants, which can all be classified as China-related hate while being of different sub-types. 
\textcolor{black}{We leverage a combination of hate speech models focusing on different aspects of hate speech to} to ensure the wide coverage of various types of hate \textcolor{black}{needed for our research purpose}. Each tweet identified as hateful by any of the three models is considered hateful in this research.

\subsection{COVID-HATE BERT Model}

The COVID-HATE BERT model is a BERT model trained on the anti-Asian hate speech dataset COVID-HATE \citep{he2021racism}. The dataset contains 3,355 English tweets manually labeled in three categories: hate speech, counterspeech, and neutral. The BERT model trained on this dataset was able to achieve an average macro F1 score of 0.832 and a per-class F1 score on hate speech of 0.762 on the COVID-HATE test data. 
\textcolor{black}{We} use the pretrained language model directly to classify the collected tweets \textcolor{black}{without preprocessing, as it is not a step suggested in the original paper \citep{he2021racism} or the code documentations.\footnote{\href{http://claws.cc.gatech.edu/covid}{http://claws.cc.gatech.edu/covid}}} \textcolor{black}{We convert the predicted labels into binary annotations of $\pm$hateful in a post-processing step. We consider neither neutral nor counterspeech labels as hateful for the purposes of this study.} Given that the model is trained on the COVID-HATE data and proved high F-score, we assume that the model predicts Covid-related hate speech sufficiently well.

\subsection{HateXplain BERT Model}

The HateXplain BERT model is a BERT model primarily trained on the HateXplain dataset \citep{mathew2020hatexplain}.\footnote{Additional data from Gab, Twitter, and Human Rationales were included for training to boost model performance.} The dataset consists of 20K posts (in English) from Gab and Twitter. Each post has been annotated with one of the \textcolor{black}{three labels:} hate, offensive, normal. Additionally, the posts contain information on target communities and rationales behind the hate speech label provided by each annotator. The dataset covers hate speech based on race, gender, religion, sexual orientation, and miscellaneous (e.g., indigenous, refugee, immigrant). The BERT model trained on this dataset was able to reach a macro F1 score of 0.68. We use this model to identify hate speech of a wide range of types with reliable performance.

Before feeding our unlabelled data to the HateXplain BERT model for prediction, each tweet was preprocessed by cleaning the URLs, emojis, and user tokens following the steps of preprocessing tweets for hate speech detection suggested in \citet{perez2021pysentimiento}. \textcolor{black}{In a post-processing step, we binarize the model predictions by considering tweets labelled as offensive not hateful.}

\subsection{Twitter RoBERTa Hate Model}

The Twitter RoBERTa Hate model is the top-performing hate speech detection model retrained in TweetEval \citep{barbieri2020tweeteval}, an evaluation framework of Twitter-specific classification tasks, on the hate speech dataset proposed in \citet{basile-etal-2019-semeval}. The model performance ranks among the top ones on the leaderboard in TweetEval. The data used for retraining the RoBERTa model is composed of non-hateful and hateful English tweets targeting immigrants and women. The original dataset proposed in SemEval-2019 Task 5 \citep{basile-etal-2019-semeval} also contains Spanish tweets.

In order to label our data using the \textcolor{black}{binary} Twitter RoBERTa Hate model, each tweet was preprocessed by replacing URLs and user mentions with placeholders following the steps suggested in \citet{barbieri2020tweeteval}.

\section{Analysis}\label{sec:analysis}

\subsection{Overview}

Among the 1,236,335 \#china tweets collected in 2020, 2.5\% (31,500) are identified as hateful \textcolor{black}{by at least one of the three models}. 
\textcolor{black}{This percentage decreased to 1.9\% (17,872) for the 935,998 tweets posted in 2021.} Both percentages are well above the average percentage of online hateful language on Twitter estimated in \citet{gao2017recognizing}.
Overall, the number of both \#china tweets and hateful \#china tweets declined in 2021 from 2020, with the hateful percentage also decreased. 

\subsection{Daily Number and Hateful Rate Analysis}

\begin{figure}[h!]
    \centering
    \includegraphics[width=8cm]{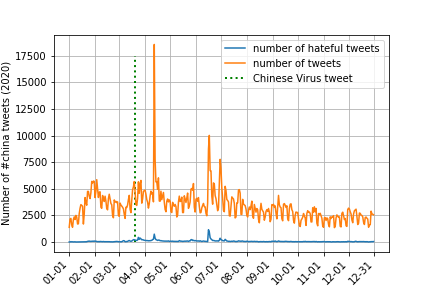}
    \caption{The number of \#china tweets and hateful \#china tweets per day in 2020.}
    \label{fig:doubleline2020}
\end{figure}

Figure \ref{fig:doubleline2020} shows the number of tweets and hateful tweets posted with the hashtag \#china per day in the year 2020.
Several visible summits of the number of daily \#china tweets can be seen in January, April, May, and June. The peaks appear mostly in the first two quarters of 2020, i.e. the beginning of the global pandemic. It is notable that in the last two quarters of 2020, the number of tweets hashtagged \#china posted per day went down to fluctuating near a lower level at 2,500, with no visible spikes. It is worth mentioning that the most outstanding spikes over the year were not triggered by the ``Chinese Virus'' tweet, instead, these happened in April and June. Both peak values surpassed 17,500 (April) and 10,000 (June) per day respectively, with no known events closely related to China that happened during the time.

\begin{figure}[h!]
    \centering
    \includegraphics[width=8cm]{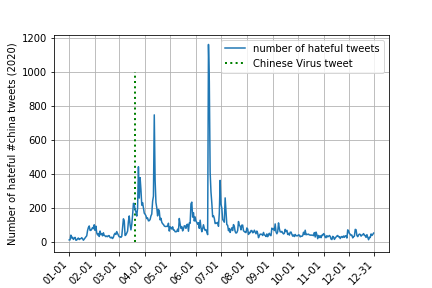}
    \caption{The number of hateful \#china tweets posted per day in 2020.}
    \label{fig:hateline2020}
\end{figure}

Figure \ref{fig:hateline2020} provides a closer look into the number of hateful \#china tweets posted each day in 2020. It is clear from the chart that the ``Chinese Virus'' tweet triggered a rise in the number of hateful tweets targeting China. However, the bigger apex values were reached in April and June. These two peaks are synced with the two most notable peaks of the daily number of \#china tweets. Overall, it can be seen from the figure that the major summits and spikes of the daily number of hateful \#china tweets were in the first two quarters of 2020. In the last two quarters, the number, in general, maintains a low level with no remarkable surges.

\begin{figure}[h!]
    \centering
    \includegraphics[width=8cm]{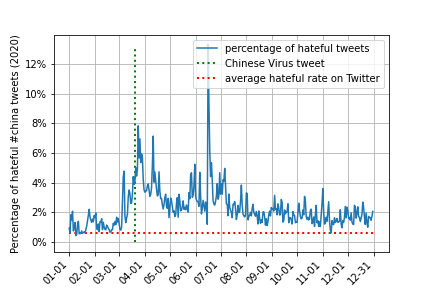}
    \caption{The daily hateful rate of \#china tweets in 2020.}
    \label{fig:perline2020}
\end{figure}

Figure \ref{fig:perline2020} shows the hateful tweet rate in \#china tweets on each day over the year 2020. It can be seen from the chart that the hateful tweet rate in \#china tweets is above the average rate on Twitter of 0.6\% most time of the year. It is notable that the hateful rate had been climbing before the ``Chinese Virus'' tweet, only the tweet further increased the rate to a higher value at 8\%, which was followed by another apex in April. The most outstanding spike appeared in June, the maximum peak value was reached at a percentage as high as over 12\%. In general, the major peaks in hateful rates are consistent with these the number of hateful tweets in 2020. Since the beginning of the pandemic in February, the hateful tweet rate of \#china tweets stays at a higher level until the end of the year with several notable surges in the second quarter.

\begin{figure}[h!]
    \centering
    \includegraphics[width=8cm]{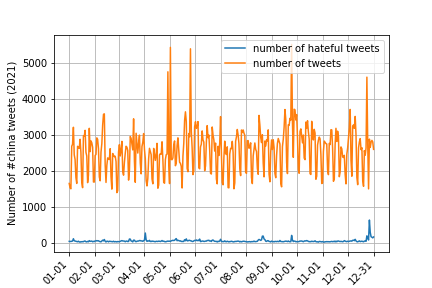}
    \caption{The number of \#china tweets and hateful \#china tweets per day in 2021.}
    \label{fig:doubleline2021}
\end{figure}

Figure \ref{fig:doubleline2021} shows the number of tweets posted with the hashtag \#china on each day in 2021, and the number of these identified as hateful per day. Apart from several apexes in April, May, October, and December, the number of \#china tweets posted per day in 2021 generally fluctuates around 2,500. The overall level of the number of tweets hashtagged \#china posted per day is lower than that in 2020. 

\begin{figure}[h!]
    \centering
    \includegraphics[width=8cm]{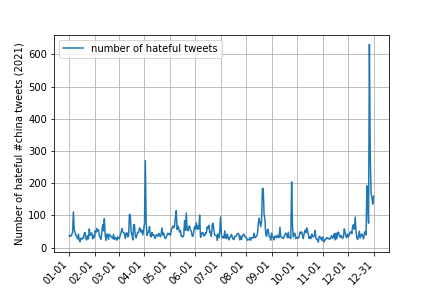}
    \caption{The number of hateful \#china tweets posted per day in 2021.}
    \label{fig:hateline2021}
\end{figure}

Figure \ref{fig:hateline2021} presents a better view over the number of \#china tweets identified hateful each day in 2021. It can be seen from the chart that the number of hateful tweets posted each day rarely surpasses 100. For most time of the year, the number maintains at a low level except for the few peaks in April, August, September, and December, with the surge in December especially outstanding, which led to the maximum apex value of over 600 hateful \#china tweets on one day in 2021.

\begin{figure}[h!]
    \centering
    \includegraphics[width=8cm]{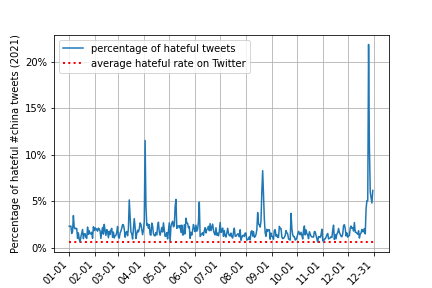}
    \caption{The daily hateful rate of \#china tweets in 2021.}
    \label{fig:perline2021}
\end{figure}

Figure \ref{fig:perline2021} shows the hateful rate in \#china tweets posted in 2021 per day. It can be seen that the several peaks in the percentage change are in harmony with the peaks in the number of hateful \#china tweets per day over the year. Apart from the smaller peak values below the level of 5\% in April, May, June, and October, the percentage reached a more outstanding apex value of over 10\% at the end of March, and around 8\% in around mid-August. It is also notable that at the end of December 2021, there was an outstanding spike that pushes the daily rate to over 20\%, which was the highest hateful rate over the entire year of 2021. Overall, throughout the entire year of 2021, the hateful rate \textcolor{black}{generally remains between} \textcolor{black}{ 0.5\% -- 3\%}
except for a few apexes. However, the rate is still overall above the average hateful rate of 0.6\% on Twitter.

\subsection{Hateful Keywords Analysis}

\begin{figure}[h!]
    \centering
    \includegraphics[width=7.5cm]{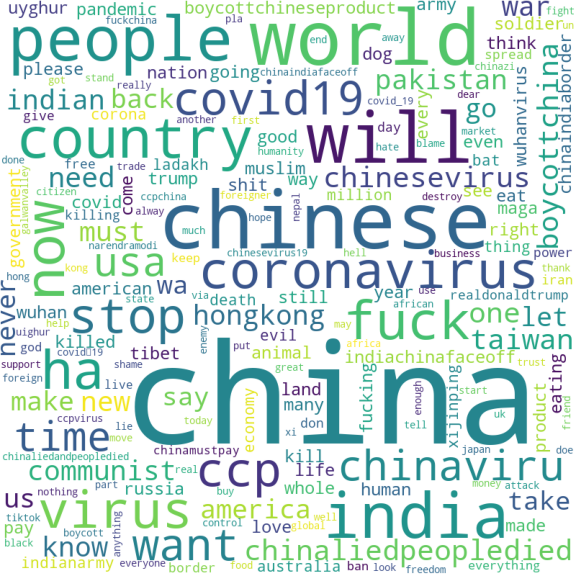}
    \caption{Word cloud generated from hateful \#china tweets posted in 2020.}
    \label{fig:cloud2020}
\end{figure}

\begin{figure}[h!]
    \centering
    \includegraphics[width=7.5cm]{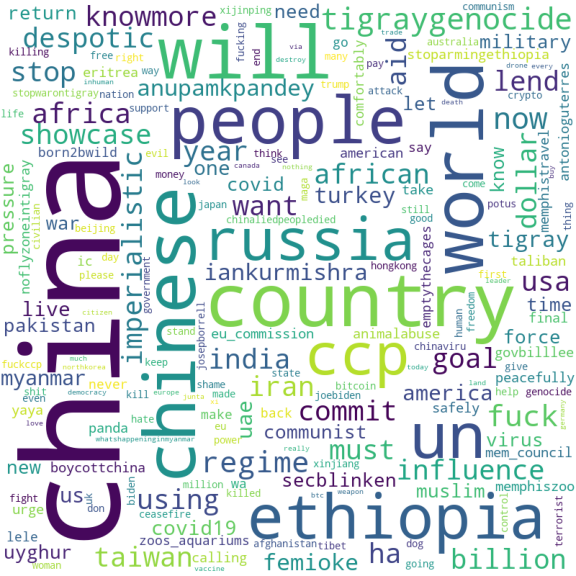}
    \caption{Word cloud generated from hateful \#china tweets posted in 2021.}
    \label{fig:cloud2021}
\end{figure}

Figure \ref{fig:cloud2020} and Figure \ref{fig:cloud2021} show the word clouds generated from the hateful tweets posted with the hashtag \#china in 2020 and 2021 respectively. We first obtained the lemma of each lower-cased token in the tweets using NLTK \citep{bird2009natural} to generate meaningful word clouds. Both figures present a visual overview of the most important and frequently mentioned keywords in the hateful \#china tweets each year. With no surprise, the keyword ``China'' is most frequently mentioned in both years as all tweets were collected according to the keyword. Interestingly, also ``US'' \textcolor{black}{and ``CCP'' (Chinese Communist Party)} are frequently mentioned keyword in the hateful \#china tweets in both years \textcolor{black}{possibly pointing to a political dimension}. \textcolor{black}{Despite these similarities, there are notable differences in the contents of hate speech in 2020 and 2021.}

It can be seen from Figure \ref{fig:cloud2020} that in 2020 some of the Covid-related terms such as ``coronavirus'', ``virus'', ``chinesevirus'', and ``covid19'' have been frequently mentioned in the \#china tweets detected as hateful. Also ``fuck''
occurs with a notable frequency in hateful tweets. Furthermore, several countries including India, Pakistan, and USA/America appear frequently in the hateful \#china tweets posted in 2020.

When it comes to the word cloud generated from hateful \#china tweets posted in 2021, it can be seen from Figure \ref{fig:cloud2021} that \textcolor{black}{the hateful discourse shifted away from the pandemic. We can identify three major topics that seem to be used to promote anti-China hate speech.}
First, the most relevant and frequently mentioned terms can be linked to Africa-related issues. We can infer from key terms such as ``Africa'', ``Ethiopia'', ``lending'', and ``dollars'' that the hate associated with \#china in 2021 was related to criticism of Chinese loans to Africa. 
\textcolor{black}{Second, we see terms associated with humanitarian crises such as the Tigray Genocide, the Uyhur population, and the coup d'état in Myanmar. Finally, several terms link to geo-political conflicts (e.g., ``Taiwan'', ``safely'', ``peacefully'', ``military'') and conflicts between different political systems (e.g., ``communist'', ``despotic'').}

\section{Conclusion}\label{sec:conclusion}

This paper presents a preliminary analysis of online hate associated with \#china on Twitter by examining hateful speech in tweets posted with the hashtag \#china over the two years following the outbreak of the Covid-19 pandemic. We collected over two million \#china tweets posted in 2020 and 2021. Through a \textcolor{black}{mutli-model approach using three different} state-of-the-art pretrained language models for hate speech detection, this study identified a wide range of hateful tweets of various hate types in these \#china tweets. \textcolor{black}{We created} a large-scale automatically labeled anti-China hate speech dataset \textcolor{black}{representing a uniquely broad range of hate speech types}. \textcolor{black}{We make our data and code available to support future work at the intersection of computational social studies and computational linguistics.} \textcolor{black}{We quantified the longitudinal development of anti-China sentiment on Twitter by tracking the daily number of \#china tweets and hateful \#china tweets, and the daily hateful rate associated with the hashtag \#china over the two-year time span. Our data and analysis allow us to quantify the longitudinal impact of events (such as Trump's ``Chinese Virus'' tweet) on the toxicity of online discourse.} \textcolor{black}{We quantified the overall level of hate associated with \#china, and the online interests with the hashtag on Twitter in both years. We observed a decrease in the number of \#china tweets, hateful \#china tweets, and hateful rate of \#china tweets from 2020 to 2021. Although in both years, the identified hateful rate of \#china tweets is above the average hateful rate of 0.6\% on Twitter \citep{gao2017recognizing}, which points towards a new direction for future research on hate speech targeting specific groups (i.e. sinophobia).} Through keyword examination, the analysis reveals the most frequently mentioned keywords in the identified hateful \#china tweets. \textcolor{black}{We observed a topic shift of general anti-China hate speech on Twitter, from Covid-related in 2020 to more diverse geo-political topics in 2021.}



\section*{Limitations}





First of all, the analysis of this research is based on hateful tweets identified through a fully automatic labeling approach using three pretrained language models.
The state-of-the-art models have been carefully chosen and proved advanced performance on hate speech detection in previous work \citep{he2021racism, mathew2020hatexplain, barbieri2020tweeteval}. \textcolor{black}{Still, it remains unclear how well these models generalize to our data which limits the impact of our findings and the usability of the data for future sociological and computational linguistic research. To address this issue, a randomly sampled subset of our dataset needs to be annotated by trained human annotators to perform a formal evaluation of our method. This would also be a valuable contribution to work on hate speech detection in general, as it goes towards validating the robustness of state-of-the-art models across data sets. It would also help to better understand the similarities and differences between the different models used in this research. This is a valuable research direction in itself because it addresses the so far under-researched question of how generalizable hate speech models are across hate speech domains. Additionally, we propose to conduct future manual annotation on a subset of the tweets identified as hateful in this research to create a gold standard \#china hate speech dataset.} 


Second, our analysis focused on visualizing \textcolor{black}{general trends in the} data. \textcolor{black}{Our interpretations of the sociological dimension of our findings have been somewhat limited and many open questions remained. It is unclear how to explain some of the peaks in hate speech observed in our longitudinal analysis.} For instance, the spike of both the number of \#china tweets and hateful \#china tweets in December 2021 remains unexplained, since no significant event known to be related to China has happened during the time. {It would be important to deepen the analysis of this data in future work to uncover the latent political and social factors leading to these} abrupt surges and outstanding peak values in our visualized data. 
\textcolor{black}{Similarly, our comparison of topics represented in hate speech in 2020 and 2021 has for now focused on a relatively simple keyword-based content analysis. In computational linguistics, more sophisticated approaches to topic modeling have been proposed in recent years \citep{lau2012line}.} A more sophisticated approach to automatic discourse modeling would likely allow us to deepen our insights into the discourse contexts in which anti-China hate speech has been introduced and aggregated. Our preliminary analysis of the data already revealed interesting trends that future work can build on.

Third, \textcolor{black}{it would be desirable to broaden the data basis of our dataset}. The analysis conducted in this research has been limited to tweets. \textcolor{black}{We for now excluded} quotes, replies and retweets. \textcolor{black}{Even though this is a common restriction in computational linguistic work on Twitter data, it clearly provides only a partial and potentially biased representation of the \#china discourse on Twitter. Quotes, replies, and retweets might include further insights relevant to the understanding of how hate speech spreads and anti-China sentiment has developed over the last two years.} Future work \textcolor{black}{should} expand the investigation scope by including these types of Twitter data. Similarly, \textcolor{black}{our analysis has so far ignored meta-information such as} user id, user networks, geo-locations, and hate type (e.g. racism) of the hateful \#china tweets. Although the meta-information of each tweet is included in our dataset, the analysis scope of our present work is limited. Future work can consider including the factors mentioned above which might yield further valuable insights into the data.


Finally, \textcolor{black}{our analysis has been limited by its focus on the English language and its time range (2020--2021). To gain a more comprehensive picture of the development of anti-China sentiment online in the wake of the Covid pandemic, it would be important to consider tweets in languages other than English. However, this extension is currently being prevented by the availability of high-performance multilingual language models for hate speech detection. Also,} examining the hate speech \textcolor{black}{level in \#china tweets before 2020} would be an important baseline to compare our findings against. \textcolor{black}{Currently, we can observe that hate speech in \#china tweets increased drastically after the first quarter of 2020 and leveled off in 2021 but without dropping to a hate speech level that we would expect based on findings from previous research. It remains unclear if this relatively high level is a residue or consequence of the anti-China sentiment promoted through the discourse around Covid. The limited data we have from the first quarter of 2020 seems to indicate that there was a relatively high level of hate speech in English \#china tweets before. On top of this, it would be worthwhile to compare online hate levels associated with different countries by inspecting the hateful rate in tweets posted with the corresponding country hashtags (e.g. \#japan, \#brazil, \#france) in future work.}





\bibliography{anthology,custom}

\begin{thebibliography}{47}
\expandafter\ifx\csname natexlab\endcsname\relax\def\natexlab#1{#1}\fi

\bibitem[{Ameur and Aliane(2021)}]{ameur2021aracovid19}
Mohamed Seghir~Hadj Ameur and Hassina Aliane. 2021.
\newblock Aracovid19-mfh: Arabic covid-19 multi-label fake news \& hate speech
  detection dataset.
\newblock \emph{Procedia Computer Science}, 189:232--241.

\bibitem[{An et~al.(2021)An, Kwak, Lee, Jun, and Ahn}]{an2021predicting}
Jisun An, Haewoon Kwak, Claire~Seungeun Lee, Bogang Jun, and Yong-Yeol Ahn.
  2021.
\newblock {Predicting anti-Asian hateful users on Twitter during COVID-19}.
\newblock \emph{arXiv preprint arXiv:2109.07296}.

\bibitem[{Badjatiya et~al.(2017)Badjatiya, Gupta, Gupta, and
  Varma}]{badjatiya2017deep}
Pinkesh Badjatiya, Shashank Gupta, Manish Gupta, and Vasudeva Varma. 2017.
\newblock Deep learning for hate speech detection in tweets.
\newblock In \emph{Proceedings of the 26th international conference on World
  Wide Web companion}, pages 759--760.

\bibitem[{Barbieri et~al.(2020)Barbieri, Camacho-Collados, Espinosa-Anke, and
  Neves}]{barbieri2020tweeteval}
Francesco Barbieri, Jose Camacho-Collados, Luis Espinosa-Anke, and Leonardo
  Neves. 2020.
\newblock {TweetEval:Unified Benchmark and Comparative Evaluation for Tweet
  Classification}.
\newblock In \emph{Proceedings of Findings of EMNLP}.

\bibitem[{Bashar et~al.(2021)Bashar, Nayak, Luong, and
  Balasubramaniam}]{bashar2021progressive}
Md~Abul Bashar, Richi Nayak, Khanh Luong, and Thirunavukarasu Balasubramaniam.
  2021.
\newblock Progressive domain adaptation for detecting hate speech on social
  media with small training set and its application to covid-19 concerned
  posts.
\newblock \emph{Social Network Analysis and Mining}, 11(1):1--18.

\bibitem[{Basile et~al.(2019)Basile, Bosco, Fersini, Nozza, Patti,
  Rangel~Pardo, Rosso, and Sanguinetti}]{basile-etal-2019-semeval}
Valerio Basile, Cristina Bosco, Elisabetta Fersini, Debora Nozza, Viviana
  Patti, Francisco~Manuel Rangel~Pardo, Paolo Rosso, and Manuela Sanguinetti.
  2019.
\newblock \href {https://doi.org/10.18653/v1/S19-2007} {{S}em{E}val-2019 {T}ask
  5: {M}ultilingual {D}etection of {H}ate {S}peech {A}gainst {I}mmigrants and
  {W}omen in {T}witter}.
\newblock In \emph{Proceedings of the 13th International Workshop on Semantic
  Evaluation}, pages 54--63, Minneapolis, Minnesota, USA. Association for
  Computational Linguistics.

\bibitem[{Bird et~al.(2009)Bird, Klein, and Loper}]{bird2009natural}
Steven Bird, Ewan Klein, and Edward Loper. 2009.
\newblock \emph{Natural language processing with Python: analyzing text with
  the natural language toolkit}.
\newblock " O'Reilly Media, Inc.".

\bibitem[{Bisht et~al.(2020)Bisht, Singh, Bhadauria, Virmani
  et~al.}]{bisht2020detection}
Akanksha Bisht, Annapurna Singh, HS~Bhadauria, Jitendra Virmani, et~al. 2020.
\newblock {Detection of hate speech and offensive language in Twitter data
  using LSTM model}.
\newblock In \emph{Recent Trends in Image and Signal Processing in Computer
  Vision}, pages 243--264. Springer.

\bibitem[{Burnap et~al.(2014)Burnap, Williams, Sloan, Rana, Housley, Edwards,
  Knight, Procter, and Voss}]{burnap2014tweeting}
Pete Burnap, Matthew~L Williams, Luke Sloan, Omer Rana, William Housley, Adam
  Edwards, Vincent Knight, Rob Procter, and Alex Voss. 2014.
\newblock {T}weeting the terror: modelling the social media reaction to the
  {Woolwich} terrorist attack.
\newblock \emph{Social Network Analysis and Mining}, 4(1):1--14.

\bibitem[{Coppersmith et~al.(2014)Coppersmith, Dredze, and
  Harman}]{coppersmith2014quantifying}
Glen Coppersmith, Mark Dredze, and Craig Harman. 2014.
\newblock Quantifying mental health signals in twitter.
\newblock In \emph{Proceedings of the workshop on computational linguistics and
  clinical psychology: From linguistic signal to clinical reality}, pages
  51--60.

\bibitem[{Corvey et~al.(2010)Corvey, Vieweg, Rood, and
  Palmer}]{corvey2010twitter}
William~J Corvey, Sarah Vieweg, Travis Rood, and Martha Palmer. 2010.
\newblock Twitter in mass emergency: what nlp can contribute.
\newblock In \emph{Proceedings of the NAACL HLT 2010 workshop on computational
  linguistics in a world of social media}, pages 23--24.

\bibitem[{Cotik et~al.(2020)Cotik, Debandi, Luque, Miguel, Moro, P{\'e}rez,
  Serrati, Zajac, and Zayat}]{cotik2020study}
Viviana Cotik, Natalia Debandi, Franco~M Luque, Paula Miguel, Agust{\'\i}n
  Moro, Juan~Manuel P{\'e}rez, Pablo Serrati, Joaquin Zajac, and Demi{\'a}n
  Zayat. 2020.
\newblock A study of hate speech in social media during the covid-19 outbreak.

\bibitem[{De~la Pena~Sarrac{\'e}n et~al.(2018)De~la Pena~Sarrac{\'e}n, Pons,
  Cuza, and Rosso}]{de2018hate}
Gretel~Liz De~la Pena~Sarrac{\'e}n, Reynaldo~Gil Pons, Carlos Enrique~Muniz
  Cuza, and Paolo Rosso. 2018.
\newblock Hate speech detection using attention-based {LSTM}.
\newblock \emph{EVALITA evaluation of NLP and speech tools for Italian},
  12:235.

\bibitem[{Devlin et~al.(2018)Devlin, Chang, Lee, and
  Toutanova}]{devlin2018bert}
Jacob Devlin, Ming-Wei Chang, Kenton Lee, and Kristina Toutanova. 2018.
\newblock {BERT}: Pre-training of deep bidirectional transformers for language
  understanding.
\newblock \emph{arXiv preprint arXiv:1810.04805}.

\bibitem[{Edelmann et~al.(2020)Edelmann, Wolff, Montagne, and
  Bail}]{edelmann2020computational}
Achim Edelmann, Tom Wolff, Danielle Montagne, and Christopher~A Bail. 2020.
\newblock Computational social science and sociology.
\newblock \emph{Annual Review of Sociology}, 46(1):61.

\bibitem[{Gao et~al.(2017)Gao, Kuppersmith, and Huang}]{gao2017recognizing}
Lei Gao, Alexis Kuppersmith, and Ruihong Huang. 2017.
\newblock Recognizing explicit and implicit hate speech using a weakly
  supervised two-path bootstrapping approach.
\newblock \emph{arXiv preprint arXiv:1710.07394}.

\bibitem[{He et~al.(2021)He, Ziems, Soni, Ramakrishnan, Yang, and
  Kumar}]{he2021racism}
Bing He, Caleb Ziems, Sandeep Soni, Naren Ramakrishnan, Diyi Yang, and Srijan
  Kumar. 2021.
\newblock {Racism is a virus: anti-Asian hate and counterspeech in social media
  during the COVID-19 crisis}.
\newblock In \emph{Proceedings of the 2021 IEEE/ACM International Conference on
  Advances in Social Networks Analysis and Mining}, pages 90--94.

\bibitem[{Hochreiter and Schmidhuber(1997)}]{hochreiter1997long}
Sepp Hochreiter and J{\"u}rgen Schmidhuber. 1997.
\newblock Long short-term memory.
\newblock \emph{Neural computation}, 9(8):1735--1780.

\bibitem[{Kim and Kesari(2021)}]{kim2021misinformation}
Jae~Yeon Kim and Aniket Kesari. 2021.
\newblock Misinformation and hate speech: The case of {Anti-Asian} hate speech
  during the {COVID-19} pandemic.
\newblock \emph{Journal of Online Trust and Safety}, 1(1).

\bibitem[{Kwok and Wang(2013)}]{kwok2013locate}
Irene Kwok and Yuzhou Wang. 2013.
\newblock {Locate the hate: Detecting tweets against blacks}.
\newblock In \emph{Twenty-seventh AAAI conference on artificial intelligence}.

\bibitem[{Lau et~al.(2012)Lau, Collier, and Baldwin}]{lau2012line}
Jey~Han Lau, Nigel Collier, and Timothy Baldwin. 2012.
\newblock On-line trend analysis with topic models:\# twitter trends detection
  topic model online.
\newblock In \emph{Proceedings of COLING 2012}, pages 1519--1534.

\bibitem[{Lazer et~al.(2009)Lazer, Brewer, Christakis, Fowler, and
  King}]{lazer2009life}
David Lazer, D~Brewer, N~Christakis, J~Fowler, and G~King. 2009.
\newblock Life in the network: the coming age of computational social.
\newblock \emph{Science}, 323(5915):721--723.

\bibitem[{Ledford(2020)}]{ledford2020facebook}
Heidi Ledford. 2020.
\newblock How facebook, twitter and other data troves are revolutionizing
  social science.
\newblock \emph{Nature}, 582(7812):328--331.

\bibitem[{Liu et~al.(2019)Liu, Ott, Goyal, Du, Joshi, Chen, Levy, Lewis,
  Zettlemoyer, and Stoyanov}]{liu2019roberta}
Yinhan Liu, Myle Ott, Naman Goyal, Jingfei Du, Mandar Joshi, Danqi Chen, Omer
  Levy, Mike Lewis, Luke Zettlemoyer, and Veselin Stoyanov. 2019.
\newblock Ro{BERT}a: A robustly optimized {BERT} pretraining approach.
\newblock \emph{arXiv preprint arXiv:1907.11692}.

\bibitem[{MacAvaney et~al.(2019)MacAvaney, Yao, Yang, Russell, Goharian, and
  Frieder}]{macavaney2019hate}
Sean MacAvaney, Hao-Ren Yao, Eugene Yang, Katina Russell, Nazli Goharian, and
  Ophir Frieder. 2019.
\newblock Hate speech detection: Challenges and solutions.
\newblock \emph{PloS one}, 14(8):e0221152.

\bibitem[{Mathew et~al.(2020)Mathew, Saha, Yimam, Biemann, Goyal, and
  Mukherjee}]{mathew2020hatexplain}
Binny Mathew, Punyajoy Saha, Seid~Muhie Yimam, Chris Biemann, Pawan Goyal, and
  Animesh Mukherjee. 2020.
\newblock {HateXplain}: A benchmark dataset for explainable hate speech
  detection.
\newblock \emph{arXiv preprint arXiv:2012.10289}.

\bibitem[{McCormick et~al.(2017)McCormick, Lee, Cesare, Shojaie, and
  Spiro}]{mccormick2017using}
Tyler~H McCormick, Hedwig Lee, Nina Cesare, Ali Shojaie, and Emma~S Spiro.
  2017.
\newblock Using twitter for demographic and social science research: tools for
  data collection and processing.
\newblock \emph{Sociological methods \& research}, 46(3):390--421.

\bibitem[{Mejova et~al.(2015)Mejova, Weber, and Macy}]{mejova2015twitter}
Yelena Mejova, Ingmar Weber, and Michael~W Macy. 2015.
\newblock \emph{Twitter: a digital socioscope}.
\newblock Cambridge University Press.

\bibitem[{Nghiem and Morstatter(2021)}]{nghiem2021stop}
Huy Nghiem and Fred Morstatter. 2021.
\newblock {"Stop Asian Hate!": Refining Detection of Anti-Asian Hate Speech
  During the COVID-19 Pandemic}.
\newblock \emph{arXiv preprint arXiv:2112.02265}.

\bibitem[{Nicola(2020)}]{nicola2020emotional}
Montemurro Nicola. 2020.
\newblock The emotional impact of covid-19: From medical staff to common
  people.
\newblock \emph{Brain, Behav., Immun}.

\bibitem[{Noble(2006)}]{noble2006support}
William~S Noble. 2006.
\newblock What is a support vector machine?
\newblock \emph{Nature biotechnology}, 24(12):1565--1567.

\bibitem[{Nockleby(2000)}]{nockleby2000internet}
John~T Nockleby. 2000.
\newblock Why {I}nternet {V}oting.
\newblock \emph{Loy. LAL Rev.}, 34:1023.

\bibitem[{Oboler et~al.(2012)Oboler, Welsh, and Cruz}]{oboler2012danger}
Andre Oboler, Kristopher Welsh, and Lito Cruz. 2012.
\newblock The danger of big data: Social media as computational social science.
\newblock \emph{First Monday}.

\bibitem[{Pereira-Kohatsu et~al.(2019)Pereira-Kohatsu, Quijano-S{\'a}nchez,
  Liberatore, and Camacho-Collados}]{pereira2019detecting}
Juan~Carlos Pereira-Kohatsu, Lara Quijano-S{\'a}nchez, Federico Liberatore, and
  Miguel Camacho-Collados. 2019.
\newblock Detecting and monitoring hate speech in twitter.
\newblock \emph{Sensors}, 19(21):4654.

\bibitem[{Potthast et~al.(2018)Potthast, Gollub, Komlossy, Schuster, Wiegmann,
  Garces~Fernandez, Hagen, and Stein}]{potthast-etal-2018-crowdsourcing}
Martin Potthast, Tim Gollub, Kristof Komlossy, Sebastian Schuster, Matti
  Wiegmann, Erika~Patricia Garces~Fernandez, Matthias Hagen, and Benno Stein.
  2018.
\newblock \href {https://aclanthology.org/C18-1127} {Crowdsourcing a large
  corpus of clickbait on {T}witter}.
\newblock In \emph{Proceedings of the 27th International Conference on
  Computational Linguistics}, pages 1498--1507, Santa Fe, New Mexico, USA.
  Association for Computational Linguistics.

\bibitem[{Putri et~al.(2020)Putri, Sriadhi, Sari, Rahmadani, and
  Hutahaean}]{putri2020comparison}
TTA Putri, S~Sriadhi, RD~Sari, R~Rahmadani, and HD~Hutahaean. 2020.
\newblock A comparison of classification algorithms for hate speech detection.
\newblock In \emph{Iop conference series: Materials science and engineering},
  volume 830, page 032006. IOP Publishing.

\bibitem[{Pérez et~al.(2021)Pérez, Giudici, and
  Luque}]{perez2021pysentimiento}
Juan~Manuel Pérez, Juan~Carlos Giudici, and Franco Luque. 2021.
\newblock \href {http://arxiv.org/abs/2106.09462} {{pysentimiento: A Python
  Toolkit for Sentiment Analysis and SocialNLP tasks}}.

\bibitem[{Ratner et~al.(2020)Ratner, Bach, Ehrenberg, Fries, Wu, and
  R{\'e}}]{ratner2020snorkel}
Alexander Ratner, Stephen~H Bach, Henry Ehrenberg, Jason Fries, Sen Wu, and
  Christopher R{\'e}. 2020.
\newblock {S}norkel: Rapid training data creation with weak supervision.
\newblock \emph{The VLDB Journal}, 29(2):709--730.

\bibitem[{Sap et~al.(2019)Sap, Card, Gabriel, Choi, and
  Smith}]{sap-etal-2019-risk}
Maarten Sap, Dallas Card, Saadia Gabriel, Yejin Choi, and Noah~A. Smith. 2019.
\newblock \href {https://doi.org/10.18653/v1/P19-1163} {The risk of racial bias
  in hate speech detection}.
\newblock In \emph{Proceedings of the 57th Annual Meeting of the Association
  for Computational Linguistics}, pages 1668--1678, Florence, Italy.
  Association for Computational Linguistics.

\bibitem[{Schmidt and Wiegand(2017)}]{schmidt-wiegand-2017-survey}
Anna Schmidt and Michael Wiegand. 2017.
\newblock \href {https://doi.org/10.18653/v1/W17-1101} {A survey on hate speech
  detection using natural language processing}.
\newblock In \emph{Proceedings of the Fifth International Workshop on Natural
  Language Processing for Social Media}, pages 1--10, Valencia, Spain.
  Association for Computational Linguistics.

\bibitem[{Shen et~al.(2022)Shen, He, Backes, Blackburn, Zannettou, and
  Zhang}]{shen2022xing}
Xinyue Shen, Xinlei He, Michael Backes, Jeremy Blackburn, Savvas Zannettou, and
  Yang Zhang. 2022.
\newblock On xing tian and the perseverance of anti-china sentiment online.
\newblock In \emph{Proceedings of the International AAAI Conference on Web and
  Social Media}, volume~16, pages 944--955.

\bibitem[{Warner and Hirschberg(2012)}]{warner2012detecting}
William Warner and Julia Hirschberg. 2012.
\newblock Detecting hate speech on the {W}orld {W}ide {W}eb.
\newblock In \emph{Proceedings of the second workshop on language in social
  media}, pages 19--26.

\bibitem[{Waseem and Hovy(2016)}]{waseem2016hateful}
Zeerak Waseem and Dirk Hovy. 2016.
\newblock Hateful symbols or hateful people? predictive features for hate
  speech detection on {T}witter.
\newblock In \emph{Proceedings of the NAACL student research workshop}, pages
  88--93.

\bibitem[{Watanabe et~al.(2018)Watanabe, Bouazizi, and Ohtsuki}]{8292838}
Hajime Watanabe, Mondher Bouazizi, and Tomoaki Ohtsuki. 2018.
\newblock \href {https://doi.org/10.1109/ACCESS.2018.2806394} {Hate speech on
  twitter: A pragmatic approach to collect hateful and offensive expressions
  and perform hate speech detection}.
\newblock \emph{IEEE Access}, 6:13825--13835.

\bibitem[{Webb et~al.(2010)Webb, Keogh, and Miikkulainen}]{webb2010naive}
Geoffrey~I Webb, Eamonn Keogh, and Risto Miikkulainen. 2010.
\newblock Na{\"\i}ve bayes.
\newblock \emph{Encyclopedia of machine learning}, 15:713--714.

\bibitem[{Wright(1995)}]{wright1995logistic}
Raymond~E Wright. 1995.
\newblock Logistic {R}egression.

\bibitem[{Wu et~al.(2022)Wu, Zhao, Lu, and Chen}]{wu2022predicting}
Xiao-Kun Wu, Tian-Fang Zhao, Lu~Lu, and Wei-Neng Chen. 2022.
\newblock Predicting the hate: A gstm model based on covid-19 hate speech
  datasets.
\newblock \emph{Information Processing \& Management}, 59(4):102998.

\end{thebibliography}
\bibliographystyle{acl_natbib}

\end{document}